\documentclass[journal=jacsat,manuscript=article]{achemso}

\usepackage{chemformula} 
\usepackage[T1]{fontenc} 
\usepackage{graphicx}%
\usepackage{multirow}%
\usepackage{amsmath,amssymb,amsfonts}%
\usepackage{amsthm}%
\usepackage{mathrsfs}%
\usepackage[title]{appendix}%
\usepackage{textcomp}%
\usepackage{manyfoot}%
\usepackage{booktabs}%
\usepackage{algorithm}%
\usepackage{algorithmicx}%
\usepackage{algpseudocode}%
\usepackage{listings}%
\usepackage{geometry}
\usepackage{subcaption}
\usepackage{lipsum}

\author{Gihan Panapitiya}
\email{gihan.panapitiya@pnnl.gov}
\author{Peiyuan Gao}
\author{Mark C. Maupin}
\author{Emily G. Saldanha}
\email{emily.saldanha@pnnl.gov}
\affiliation[Unknown University]
{Pacific Northwest National Laboratory, Richland, WA, USA}

\title[An \textsf{achemso} demo]
  {FragNet: A Graph Neural Network for Molecular Property Prediction with Four Levels of Interpretability}

\abbreviations{IR,NMR,UV}

\begin{document}

\begin{abstract}
Molecular property prediction is essential in a variety of contemporary scientific fields, such as drug development and designing energy storage materials. Although there are many machine learning models available for this purpose, those that achieve high accuracy while also offering interpretability of predictions are uncommon. We present a graph neural network that not only matches the prediction accuracies of leading models but also provides insights on four levels of molecular substructures. This model helps identify which atoms, bonds, molecular fragments, and connections between fragments are significant for predicting a specific molecular property. Understanding the importance of connections between fragments is particularly valuable for molecules with substructures that do not connect through standard bonds. The model additionally can quantify the impact of specific fragments on the prediction, allowing the identification of fragments that may improve or degrade a property value. These interpretable features are essential for deriving scientific insights from the model's learned relationships between molecular structures and properties.
\end{abstract}

\section{Introduction}\label{sec1}

Molecular property prediction is a pivotal aspect of material discovery and design in a wide range of modern scientific fields, including drug design, energy storage material discovery, catalysis, and agrochemicals. Despite the availability of numerous machine learning models \cite{rong2020self, butler2018machine, gilmer2017neural}, there is often a trade-off between achieving highly accurate predictions and ensuring the interpretability of these predictions, which is crucial for scientific insights. The ability to interpret machine learning models is particularly important in the context of molecular property prediction because it allows scientists to understand the underlying factors that drive these predictions \cite{schutt2024schnet4aim, wang2024lamole}. This understanding can lead to the identification of new hypotheses, the discovery of novel materials with desired properties, and the acceleration of the molecular design process. Furthermore, interpretability is also helpful to understand the strengths and limitations of models and thereby develop improvements \cite{hoedl2023explainability}.

Owing to these benefits, a surge of interest has sparked in explainable Artificial Intelligence (XAI), which aims to provide a deeper understanding of the model's prediction process \cite{adadi2018peeking, rudin2019stop}. The terms explainability and interpretability have been used interchangeably in the literature to define explainable AI. Biran and Cotton \cite{biran2017explanation} define the interpretability of a model as ``the degree to which an observer can understand the cause of a decision.'' Nigam Shah \cite{shah2019importance} categorizes interpretability into three types: engineers’ interpretability, causal interpretability, and trust-inducing interpretability. Engineers’ interpretability focuses on understanding how the AI model arrived at its output or how the model functions internally. It examines the technical aspects of model architecture, algorithms, and computational processes. This approach helps developers and researchers understand data flow and transformations within the model, enabling them to debug, optimize, and refine AI systems. Causal interpretability explains why AI models make specific decisions or predictions. It identifies key factors influencing the model's output and explores causal relationships between these factors and the results. This approach provides intuitive explanations accessible to both technical and non-technical stakeholders, helping build trust in AI systems. Trust-inducing interpretability is concerned with fostering confidence in the model’s predictions. This approach aims to present the model's decision-making process in a way that is understandable and relatable to users, regardless of their technical expertise. It involves creating clear, accessible explanations of how the model arrives at its conclusions, often using simplified language or visual representations. In essence, an explainable or interpretable model is one that can provide the user with some kind of understanding regarding its underlying decision-making process.

Interpretable models can be categorized in several ways. One categorization identifies models as intrinsic or extrinsic (post-hoc), based on whether interpretability is integrated into the model’s prediction algorithm or not \cite{doshi2017towards, lundberg2017unified}. Extrinsic interpretability models offer the advantage of being applicable to any complex model, regardless of its architecture \cite{D1SC05259D}. They provide valuable insights into the behavior of black-box models without the need to alter the models themselves, making them particularly useful for validating complex algorithms and increasing trust in their outputs \cite{lipton2018mythos}. Interpretation methods such as SHAP (SHapley Additive exPlanations) \cite{lundberg2017unified} and LIME (Local Interpretable Model-agnostic Explanations) \cite{ribeiro2016should} are widely used for explaining traditional machine learning models by attributing feature importance scores. SHAP uses Shapley values from cooperative game theory to compute the contribution of each feature, while LIME approximates the model's behavior around a specific prediction with an interpretable surrogate model. While powerful for typical tabular and image data, these methods face challenges when applied to graph-based molecular data. Molecules represented as graphs contain interconnected atoms and bonds, making it difficult for SHAP and LIME to capture hierarchical relationships and specialized interactions unique to molecular properties.

Intrinsic interpretability is often more advantageous in domains where transparency and simplicity are paramount. Intrinsic models are inherently easier to understand and explain. Their transparent decision-making processes make them particularly appealing in critical areas such as healthcare and finance, where the ability to directly interpret and trust the model's decisions is essential \cite{rudin2019stop}. Intrinsic interpretable models have been developed for models that use different molecular representations \cite{rymarczyk2022progrets}. These representations include SMILES strings, molecular fingerprints, and descriptors, each offering different levels of granularity and interpretability. Feature importance values or attention weights on SMILES tokens either do not provide a comprehensive understanding of the importance of specific structural elements of the molecule or are difficult to interpret. Additionally, creating a unified interpretability functionality for different molecular representations (such as SMILES encoding, molecular descriptors, etc.) is challenging. Molecules are inherently graph-like structures, with atoms as nodes and bonds as edges. This alignment makes Graph Neural Networks (GNNs) a natural representation for molecules, allowing them to effectively capture the spatial relationships and connectivity patterns within molecules \cite{gilmer2017neural, yang2019analyzing}.

Previous work on interpretable GNNs for molecular property prediction has led to exciting developments, such as models designed to highlight important substructures or provide attention maps indicating which parts of the molecule contribute most to the prediction \cite{ying2019gnnexplainer, pope2019explainability}. The AttentiveFP \cite{xiong2019pushing} and MoGAT \cite{tanvir2024mogat} models provided atom-level importance values. The work by Wu et al. \cite{wu_chemistry-intuitive_2023} provided fragment-level attention. However, their work is limited in the range of molecular properties on which it evaluates, as it only provides results for the ESOL dataset \cite{ESOL} using a random train-test split configuration.

Despite these advancements, there remain significant limitations. Current GNN models do not take into account the interactions between molecular substructures that are not connected by covalent bonds, such as salts and complexes. Such compounds play an important role in energy and pharmaceutical applications. Furthermore, to the best of our knowledge, there is no interpretable GNN model that takes into account atom, bond, and molecular fragments within a single framework when providing interpretations. Most models either focus on atoms or molecular fragments \cite{numeroso2021meg}.

In this work, we introduce a novel graph neural network architecture named FragNet. FragNet not only achieves prediction accuracies comparable to those of state-of-the-art models, but it  also provides insightful analysis across four levels of molecular substructures: atoms, bonds, fragments, and fragment connections. Here, a `fragment' refers to a distinct substructure within a molecule, comprising groups of atoms and the covalent bonds that connect them. FragNet's architecture includes graphs for each type of substructure, thereby enabling an understanding of which atoms, bonds, molecular fragments, and connections between fragments are critical in predicting a given molecular property. By constructing a fragment-based graph, we facilitate message passing between fragments that are not directly connected by covalent bonds. To the best of our knowledge, FragNet is the first model capable of simultaneously attending to all these substructures within a single framework.

We have also implemented a method to quantify the contribution of a molecular fragment to the prediction. We show that our results on the attention values and contribution values align with the well established chemical observations, thus improving the trust of the model’s prediction process. By enhancing the model with the ability to reason about the connections between fragments, we provide a more complete representation for molecules with substructures that are not connected with regular covalent bonds, such as salts and complexes.


We demonstrate the utility of the multi-layered interpretability mechanisms of the model through several case studies on a selection of property prediction tasks. We use model attention weights and contribution values to investigate the reasoning used by the model for both individual molecular predictions as well as aggregated across multiple predictions, to identify the key molecular constituents that drive property variations. To further validate the reasoning extracted from the models, we perform a comparative study of FragNet contribution scores with density functional theory (DFT) computations of electrostatic surface potentials. Finally, we develop and release a interactive browser application to make these types of interpretability studies accessible for other molecular property tasks.

For readers who may not be familiar with the molecular chemistry concepts discussed in this work, we have provided a brief overview in the Supporting Information section.

\section{Method}\label{sec2}

\subsection{Model}

\begin{figure*}[!thb]
    \centering
    \includegraphics[width=1\textwidth]{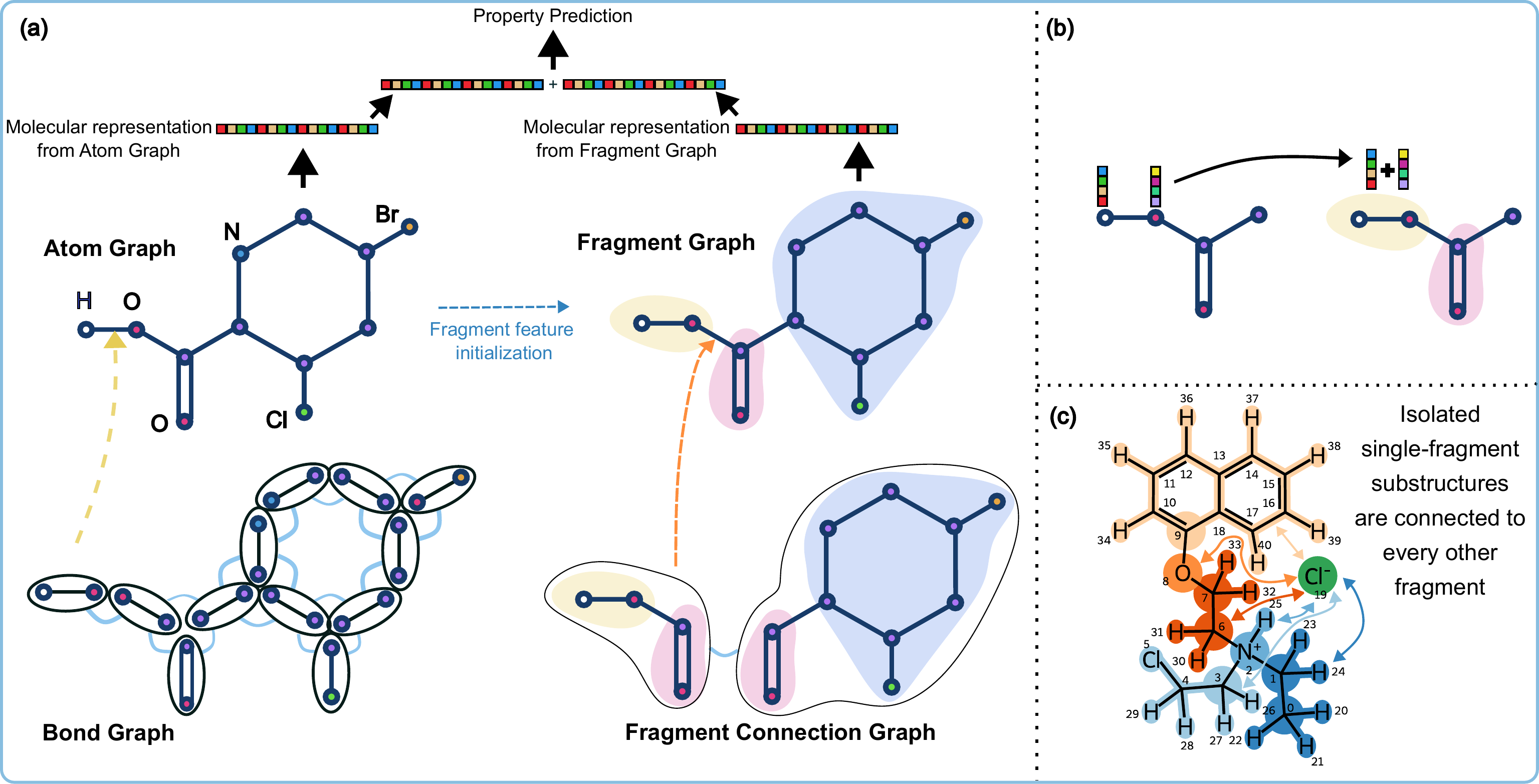}  
    \caption{FragNet's architecture and data representation. (a) Atom and Fragment graphs' edge features are learned from Bond and Fragment connection graphs respectively. b) Initial fragment features for the fragment graph are the summation of the updated atom features that compose the fragment. (c) Illustration of FragNet's message passing taking place between two non-covalently bonded substructures. Fragment-Fragment connections are also present between adjacent fragments in each non-covalently bonded structure of the compound. }
    \label{fig:fragnet_arch}
\end{figure*}

The FragNet model is based on a message-passing graph neural network architecture which reasons over four different graph-based representations of the molecular structure. An overview of the graph-based representations used by our model is shown in Figure \ref{fig:fragnet_arch}. The data representation used by our model consists of four graph structures: atom-based, bond-based, fragment-based and fragment connection-based. The nodes and edges of the atom graph are atoms and covalent bonds respectively. In the bond graph, the nodes are bonds, while an edge in this graph is formed when two bonds have a common atom. The fragment graph is created by decomposing the molecules into substructures using a fragmentation scheme. In this work, we used BRICS fragmentation\cite{Degen_BRICS, Seo_2023, Jinsong2024}. If the molecule is not a salt or a complex, the edges in the fragment graph are formed by bonds at which the fragments are split. 

In the case of molecular complexes, where regular bonds do not exist between some fragments, we form `virtual' connections to enable message passing between such disconnected substructures. Specifically, we create virtual connections between fragments that are linked by BRICS fragments.
These virtual connections can represent non-covalent interactions such as van der Waals forces, hydrogen bonds and electrostatic interactions. These interactions, while not as permanent as covalent bonds, are valid and significant for molecular modeling. If a molecule cannot be decomposed into fragments, we treat the molecule itself as a fragment and establish a self-connection. For molecules composed of substructures that are not connected by covalent bonds, we create virtual connections by linking each fragment in one substructure to each fragment in the other.

Finally, the fragment-connection graph represents the connections between fragments as nodes with an edge formed when the connections have a fragment in common. These connections can be regular and/or virtual bonds depending on the type of the molecule. An important benefit of the fragment-connection graph is that it gives the ability to understand the interactions between different substructures of the molecules. This is particularly useful when regular bonds do not exist between two substructures.

The FragNet model leverages a hierarchical approach among these graph-representations, using the learned representations of lower-level structures to initialize the features of the high-level structures in the each subsequent graph representation.  
Our model starts by creating the representation for a given molecule using the bond graph. The nodes in the bond graph are initially featurized using bond properties, while the edges in the bond graph are featurized using bond angles. As bond properties we considered  bond type (single, double, triple, aromatic), whether the bond is considered to be conjugated, whether the bond is in a ring and stereochemistry of the bond.

The node representation of the bond graph is updated using the graph attention mechanism \cite{GAT_Petar}. The updated node features of the bond graph, $h_{i}^{bond-graph}$,  are calculated as,

\begin{equation}
  h_{i}^{bond-graph} = \sum_{j \in N(i)} \alpha_{i,j}^{bond-graph}\textbf{W}h_{j,0}^{bond-graph}.
  \label{eq2}
\end{equation}

Here, $\alpha_{i,j}^{bond-graph}$ represents the importance of node j and the edge connecting nodes $i$ and $j$ to node $i$.

\begin{equation}
  \alpha_{i,j}^{bond-graph} = \frac{exp(LeakyRelU( M_{i,j} ))}{\sum_{k \in N(i)} exp(LeakyRelU( M_{i,k} ))},
  \label{eq1}
\end{equation}

where,
\begin{equation}
M_{i,j}  = \mathbf{a}^{b}[ \mathbf{W}^{b}h_{i}^{b}|| \mathbf{W}^{b}h_{j}^{b}||\mathbf{W}^{be}e_{i,j}| 
  \label{eq1.1}.
\end{equation}

$h_{j}^{b}$ is a feature vector of a neighboring node of node $i$ in the bond graph. 
Note that these nodes correspond to bonds in the atom graph. \textbf{$a^{b}$} is a trainable weight vector. $e_{i,j}$ is the  feature vector for the edge connecting nodes $i$ and $j$ in the bond graph.  These attention weights determine the level of impact that each other bond has $h_{j}^{b}$ in updating the learned representation of a given bond, $h_{i}^{b}$ .

We do not consider self edges in the bond graph, as it can significantly increase the computational cost. Hence $N(i)$ includes only the first nearest neighbors.

The updated node features of the bond graph are then used as the initial edge features of the atom graph, as indicated by the yellow arrow in Figure \ref{fig:fragnet_arch}(a). Using these edge features, the atom graph is updated, also using the graph attention mechanism. The initial node features of the atom graph are atomic number,
implicit valence, formal charge, number of radical electrons, hybridization, is the atom aromatic, is the atom
in a ring, total number of Hydrogen atoms and chirality. 

For the message passing in the atom graph, we do consider self edges (the edges connecting an atom to itself). The edge properties of the self edges are initialized as a vector of zeros having the same dimension as $h_{i}^{bond-graph}$.

\begin{equation}
  h_{i}^{atom-graph} = \sum_{j \in N(i)} \alpha_{i,j}^{atom-graph}\textbf{W}h_{j,0}^{atom-graph}
\end{equation}

Once updated, a molecular representation is created by summing all the atom feature vectors of the atom graph.

The fragment graph is updated following a similar procedure. For the initial node features of the fragment graph, we use the summed atom features from the atom graph corresponding to the atoms of which the fragment is composed. 

\begin{equation}
  h_{i,0}^{fragment-graph} = \sum_{j \in Atoms\_of\_frag_{i}} h_{j}^{atom-graph}
\end{equation}

The fragment graph's node representations are then updated through the graph attention mechanism. Then the edge features of the fragment graph are initialized using the learned node features of the fragment connection graph before applying graph attention. Using the updated fragment graph a second molecular representation is constructed. The final molecular representation is the concatenation of the representations constructed based on the atom and fragment graphs. This representation can then be used for downstream tasks like molecular property prediction. A complete mathematical description of the FragNet architecture is provided in the Methods section.

The details on the data and the training approach are provided in the Supporting Information.

\section{Results}\label{sec3}
\subsection{Prediction Accuracies}
We test the performance of FragNet on multiple molecular property prediction tasks derived from the MoleculeNet benchmark including four regression tasks and three classification tasks. These tasks spanned a range of chemical, biological, and toxicity properties. Before performing property prediction training, the model was pre-trained on a set of self-supervised tasks as described in the Methods section.
Table \ref{table:res-reg} and Table \ref{table:res-clsf} presents the prediction accuracies of FragNet obtained based on scaffold splitting\cite{Ramsundar-et-al-2019}. For each dataset, FragNet's hyperparameters were optimized based on the validation accuracy using three random seeds. Using the optimized parameters and the same three random seeds, FragNet was fine-tuned, and the mean and standard deviation of the prediction accuracy on the test set are reported in Tables \ref{table:res-reg} and \ref{table:res-clsf}. We compare the performance of FragNet with four state-of-the-art baseline approaches. With the exception of CEP and Malaria, FragNet achieves accuracies comparable to or better than the current state-of-the-art for both regression and classification tasks. For CEP, the slightly lower accuracies may be attributed to the fact that we only use the first hyperparameter combination suggested by Optuna  compared to thirty hyperparameter optimization runs  conducted for the other datasets. This is due to the time constraints of training as the training set of CEP is large, with 23,982 molecules.

\begin{table*}[!thb]
\small
\caption{FragNet performance measured with RMSE on regression tasks in the MoleculeNet benchmark compared with four state-of-the-art baselines. Best results in bold. Lower is better. Reported here are \textit{mean} $\pm$ \textit{standard deviation } corresponding to three random seeds. Except for SimSGT, results for other models were obtained from the work of reference \cite{molebert2023}.  } 
\centering
\begin{tabular}{lcccc}
\toprule

Dataset              &  ESOL & LIPO & CEP  \\
\midrule
ContextPred\cite{hu2020pretraining, molebert2023}  &  1.196 $\pm$ 0.037  &  0.702 $\pm$ 0.020  & 1.243 $\pm$ 0.025 \\
AttrMask\cite{hu2020pretraining, molebert2023}      &  1.112 $\pm$ 0.048  &  0.730 $\pm$ 0.004 & 1.256 $\pm$ 0.000\\
JOAO\cite{You2021,molebert2023}   & 1.120 $\pm$ 0.019 &  0.708 $\pm$ 0.007  & 1.293 $\pm$ 0.003  \\
GraphMVP\cite{liu2022pretraining, molebert2023}   & 1.064 $\pm$ 0.045 &  0.691 $\pm$ 0.013  & 1.2228 $\pm$ 0.001  \\
Mole-BERT\cite{molebert2023}   & 1.015 $\pm$ 0.030 &  0.676 $\pm$ 0.017  & 1.232 $\pm$ 0.009  \\

SimSGT\cite{liu2023rethinking} & 0.917 $\pm$ 0.028 &   \textbf{0.670 $\pm$ 0.015}     &  \textbf{1.036 $\pm$ 0.022} \\
FragNet &  \textbf{0.881 $\pm$ 0.011}  & 0.682 $\pm$ 0.031 & 1.092 $\pm$ 0.031\\

\bottomrule
\end{tabular}
\label{table:res-reg}
\end{table*}

\begin{table*}[!thb]
\small
\caption{FragNet performance measured with \textit{mean} $\pm$ \textit{standard deviation } AUC-ROC corresponding to three random seeds on classification tasks in the MoleculeNet benchmark compared with four state-of-the-art baselines. Best result in bold. Higher is better. }
\centering
\begin{tabular}{lcccc}
\toprule

Dataset           & Clintox & Sider & Tox21 \\
\midrule
ContextPred\cite{molebert2023}   & 74.0 $\pm$ 3.4 & 59.7  $\pm$ 1.8  & 73.6 $\pm$ 0.3\\

JOAO\cite{You2021,molebert2023} & 66.6 $\pm$ 3.1 & 60.4 $\pm$ 1.5 & 74.8 $\pm$ 0.6 \\
AttrMask\cite{hu2020pretraining,molebert2023} & 73.5 $\pm$ 4.3 & 60.5 $\pm$ 0.9  & 75.1 $\pm$ 0.9 \\
MGSSL\cite{ZHANG2021,molebert2023} & 77.1 $\pm$ 4.5 & 61.6 $\pm$ 1.0  & 75.2 $\pm$ 0.6 \\

GraphMVP\cite{liu2022pretraining,molebert2023} & 79.1 $\pm$ 2.8 & 60.2 $\pm$ 1.1  & 74.9 $\pm$ 0.8 \\
Mole-BERT\cite{molebert2023}  &  78.9 $\pm$ 3.0 & 62.8 $\pm$ 1.1   & 76.8 $\pm$ 0.5\\
SimSGT\cite{liu2023rethinking} &  85.7 $\pm$ 1.8 & 61.7 $\pm$ 0.8  & 76.8 $\pm$ 0.9\\
FragNet   & \textbf{86.8 $\pm$ 1.8}  & \textbf{63.7 $\pm$ 1.9}  & \textbf{76.9 $\pm$ 0.6}\\

\bottomrule
\end{tabular}
\label{table:res-clsf}
\end{table*}

\begin{figure*}
    \centering
    \includegraphics[width=1\textwidth]{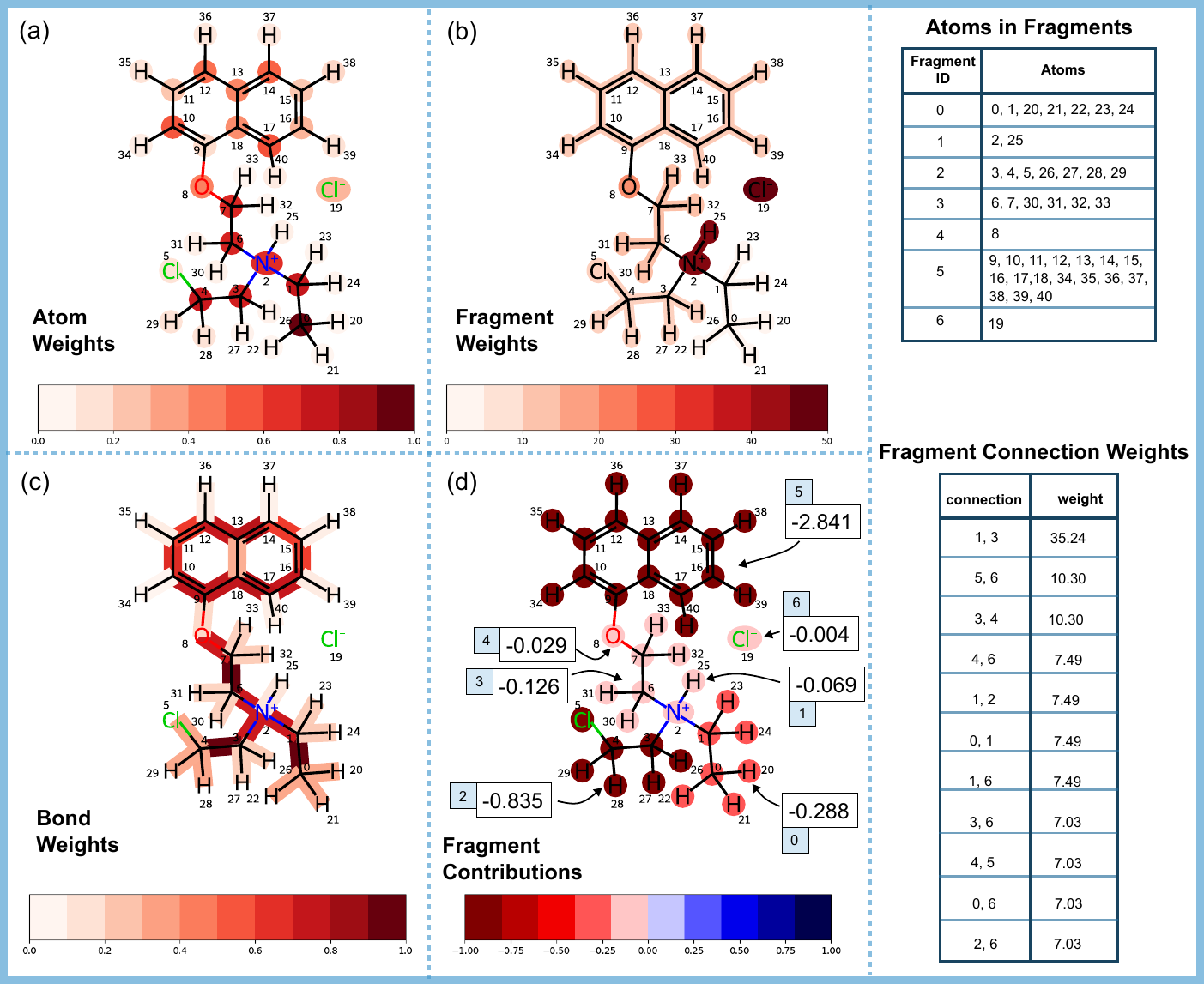}  
    \caption{Different types of attention weights and contribution values available in FragNet visualized for CC[NH+](CCCl)CCOc1cccc2ccccc12.[Cl-] with atom, bond, and fragment attention weights shown in (a),(b), and (c) and fragment contribution values shown in (d). The top table provides the atom to fragment mapping and the bottom table provides the fragment connection attention weights. Atom and bond attention weights are scaled to values between 0 and 1. The fragment and fragment connection weights are not scaled. The numbers in blue boxes in (d) correspond to Fragment IDs in `Atoms in Fragments' table.}
    \label{fig:weights}
\end{figure*}

\subsection{Interpretability}


The primary distinction between FragNet and other models lies in its ability to interpret predictions based on four different types of substructures. Notably, FragNet can handle molecules with substructures that are not connected via covalent bonds, such as salts and complexes. We can analyze all four substructure types using two different mechanisms from the model - attention weights and contribution values. Attention weights are derived from the graph attention mechanisms from each of the four graph representations and provide insight into which substructures the model focuses on when making its predictions. 
Meanwhile, we can quantify the contribution of a given substructure to the prediction of a molecular property value~\cite{wu_chemistry-intuitive_2023}. This is achieved by first predicting the property for the entire molecule in the usual manner. Subsequently, another prediction is made after masking the substructure of interest, meaning the node features of this substructure are excluded from the prediction process. If the prediction with the masked substructure is lower than that of the unmasked molecule, the masked substructure is considered property-increasing. Conversely, if the prediction with the masked substructure is higher, the masked substructure is deemed to reduce the property value. Formally, the contribution is defined as follows:

\begin{equation}
Contribution = Property^{unmasked} - Property^{masked}
\end{equation}

\begin{figure}[t]
   \centering
   \includegraphics[width=.5\textwidth]{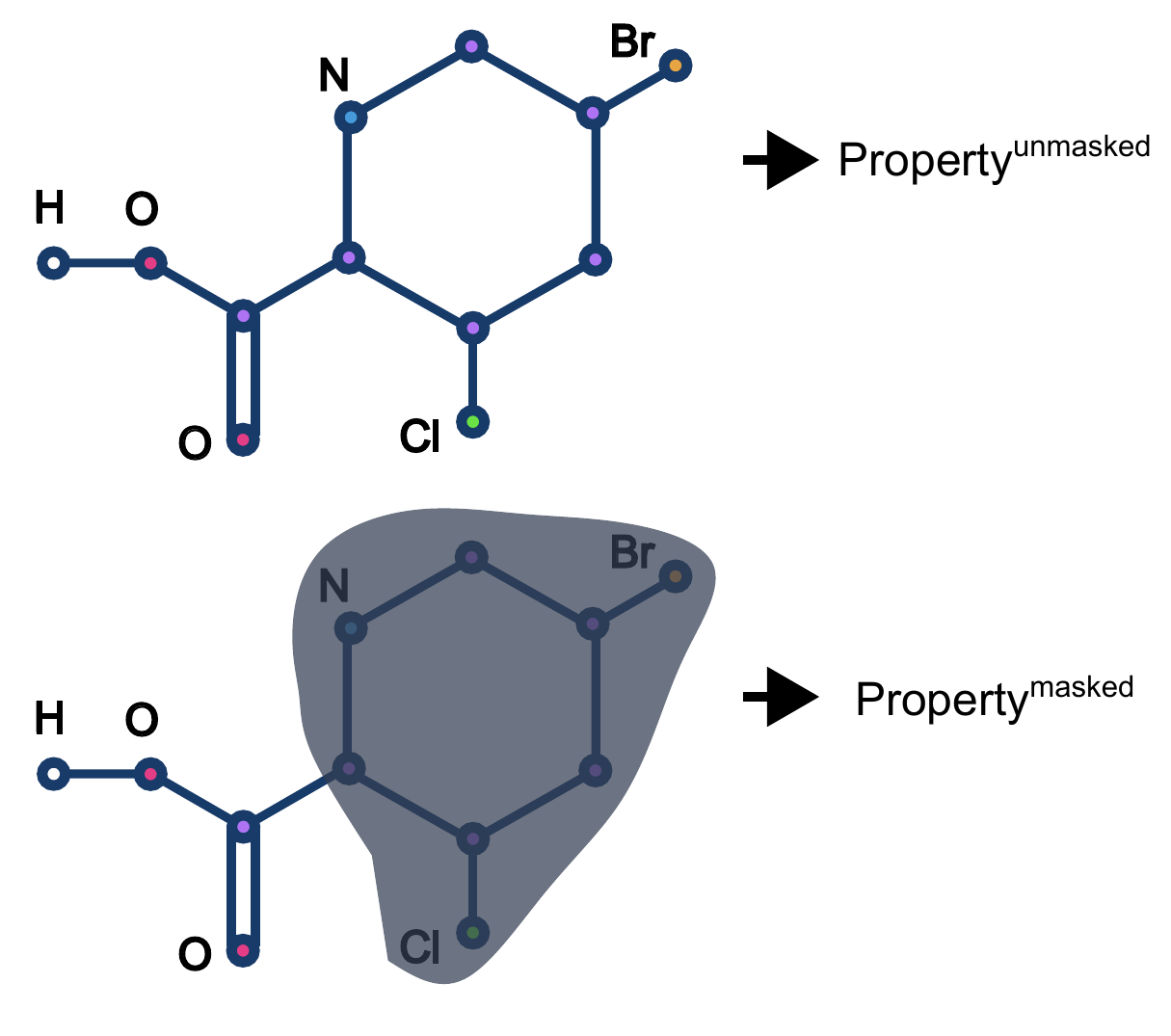}  
   \caption{Illustration of the contribution value calculation}
   \label{fig:contributions}
\end{figure}

\subsubsection{Interpretation Case Study}

We first demonstrate the interpretation capabilities of FragNet  using a sample compound. In the subsequent sections, we analyse the insights that can be gained by using attention weights and contribution values using three prediction tasks: solubility, lipophilicity and cancer drug response. 
In the following discussions, we use the notation A(\textit{i}) to denote an atom of type A with atom index \textit{i}, and A(\textit{i})-B(\textit{j}) to denote a bond connecting atoms A(\textit{i}) and B(\textit{j}).

The attention weights and contribution values of 1-Naphthyloxyethylethyl-beta-chloroethylamine hydrochloride from the solubility prediction model are illustrated in Figure \ref{fig:weights}. The largest atom attention weights are on the atoms in the vicinity of positively charged nitrogen, N(2) atom. This is expected as this region likely interacts most significantly with the solvent water molecules. The next largest weights are on four atoms in the aromatic ring and ether O(8). A significant attention on ether oxygen is also understandable as it can act as a hydrogen bond acceptor allowing it to interact with water molecules. These ether oxygen can also withdraw electrons from the adjacent -C\textsubscript{2}H\textsubscript{2}- group and donate them to double ring system\cite{Harrold2012-gr}.

In terms of bond attention weights, the largest weights are on the C(0)-C(1), C(3)-C(4), and C(6)-C(7) bonds. One atom of each of these bonds are connected to N(2)\textsuperscript{+}. Note that these three bonds are part of the fragments 0, 2, and 3. Examining the 3D structure of this molecule, we observe that these three fragments could cause steric hindrance to N(2)\textsuperscript{+}. 
We hypothesize that the large weights on C(0)-C(1) and C(3)-C(4) could be attributed to two factors: (a) the steric effects caused by fragments 0, 2, and 3, and (b) the electron flow towards the N(2)\textsuperscript{+} atom, as alkyl groups generally exhibit electron-donating properties. It should also be noted that due to the presence of chlorine in fragment 2, its electron-donating capability might be less than that of fragment 0.
Moreover, there could be an electron flow through C(6)-C(7) towards the double ring system because of the ether oxygen bonded to the fragment 3. Ethers are known to exhibit both electron-withdrawing and electron-donating effects\cite{Harrold2012-gr}. In this molecule, it is possible that the ether oxygen is withdrawing electrons from fragment 3 and donating them to the double ring system. The large weight on the C(6)-C(7) bond could signify this interesting dynamic in electron flow, as it is situated between two electron-withdrawing atoms, O(8) and N(2).



The fragments containing N\textsuperscript{+} and Cl\textsuperscript{-} are the ones with the highest attention weights. This result is not surprising, as these ionic species tend to interact strongly with solvent water molecules.
The strongest connection weight is observed between the N\textsuperscript{+}$-$H  fragment and the $-$C\textsubscript{2}H\textsubscript{4}$-$ group adjacent to the ether oxygen.
Although the exact reasons for this interaction remain unclear, it is possible that the $-$C\textsubscript{2}H\textsubscript{4}$-$ group contributes to the stability of the compound by shielding the N\textsuperscript{+} atom from the electron-withdrawing effects of the ether oxygen. Additionally, Cl\textsuperscript{-} shows significant interaction with the double ring system, while the ether oxygen interacts robustly with its adjacent hydrocarbon group. Subsequent large interaction weights are noted between the ether oxygen and Cl\textsuperscript{-}, N\textsuperscript{+} and its adjacent chloroethyl group, N\textsuperscript{+} and its adjacent ethyl group as well as between N\textsuperscript{+} and Cl\textsuperscript{-}.


In Figure~\ref{fig:weights}(d), we show the contribution values for the molecule fragments which provide information about the directional impact of different substructures on the predicted property. From the fragment contributions, we identify that the double ring system is the most hydrophobic based on its highly negative contribution value, indicating that it significantly decreases the model's predicted solubility. The next most hydrophobic groups are fragments 2 and 0. The hydrophobicity of these two fragments can be mainly attributed to the steric effects they impose on N\textsuperscript{+}. 
Given their ability to form hydrogen bonds with solvent water molecules, it is not surprising that the most hydrophilic substructures are Cl\textsuperscript{-} and those containing oxygen and N\textsuperscript{+}.


\subsubsection{Substructure Analysis}
\label{substructure_analysis}

In the following sections, we perform a quantitative study of the highly impactful substructures for predicting solubility, lipophilicity, and cancer drug response. To identify these substructures, we first filter the model predictions to those with low error to focus on cases where the model is correctly inferring the structure-property relationships. This filtration is important to ensure that the attention weights are reliable in interpreting the model predictions. We then identify substructures which appear in multiple molecules and average their attention weights across all the predictions. Substructures which have high average attention across the predictions are likely to have a significant impact on the predicted property. 

\begin{figure}
    \centering
        \centering
        \includegraphics[width=.8\textwidth]{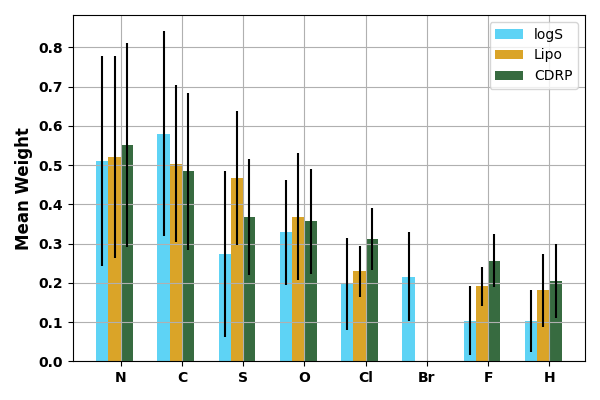}
        \caption{Mean atom attention weight values when compounds with prediction error $<$ 0.1 were considered. Error bars indicate the standard deviation. }
        \label{fig:atom-weights-0.1}
\end{figure}

\begin{figure}
    \centering
        \includegraphics[width=.8\textwidth]{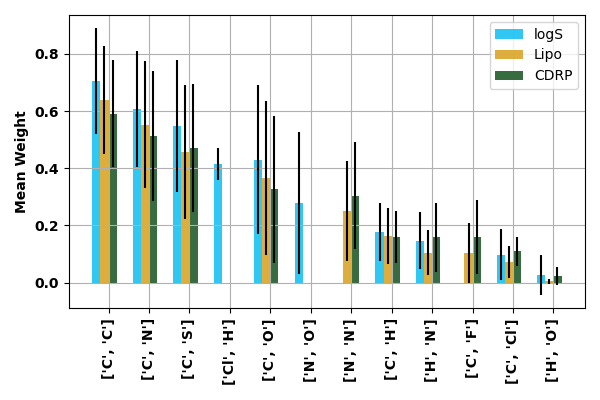}
        \caption{Mean bond attention weight values when compounds with prediction error $<$ 0.1 were considered.Error bars indicate the standard deviation.}

        \label{fig:bond-weights-0.1}
\end{figure}

\begin{figure*}[t]
    \centering

        \includegraphics[width=1\textwidth]{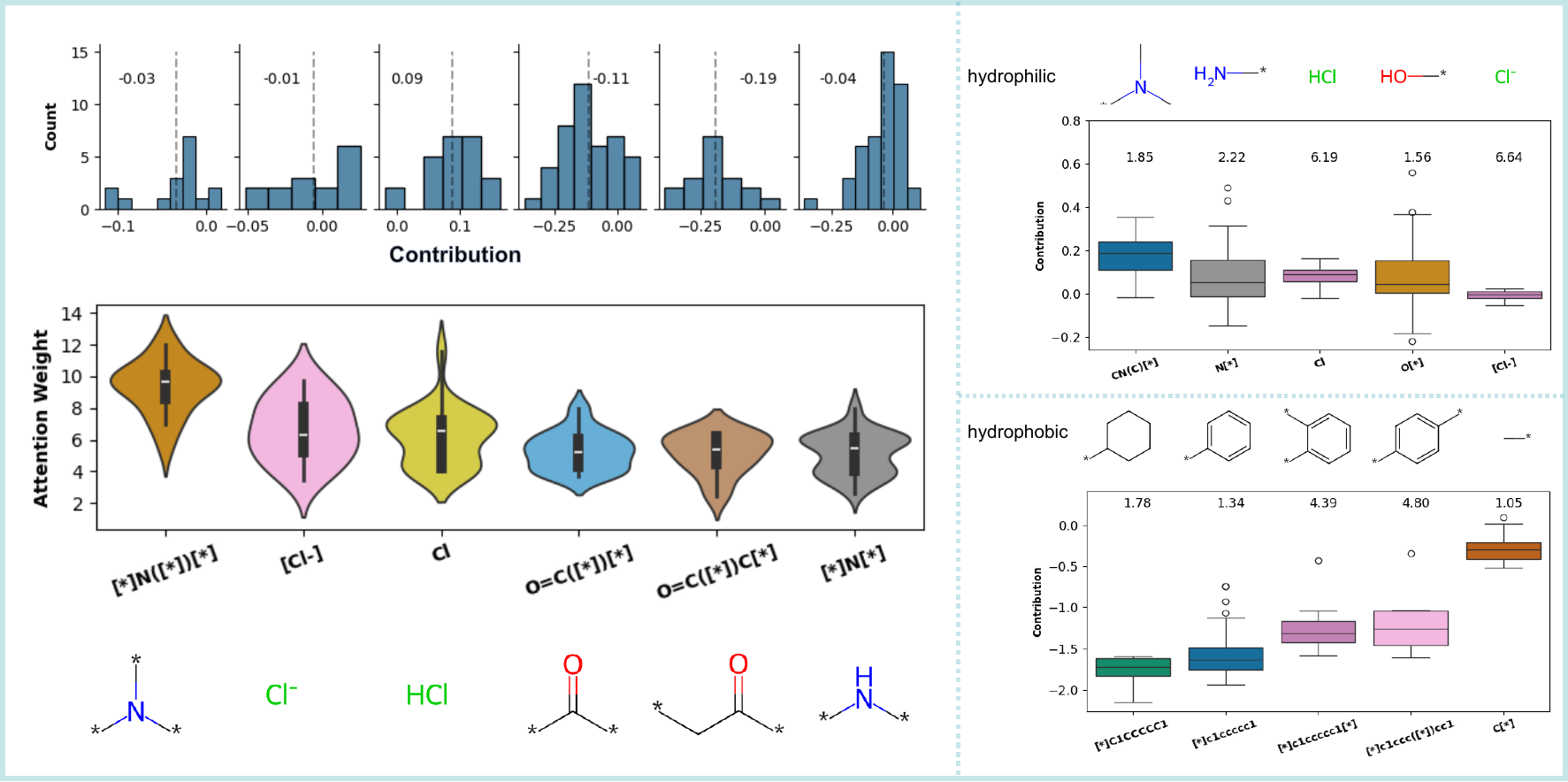}
        \caption{Fragment and fragment-fragment connection attention weights and contribution values corresponding to solubility prediction. Left: violin plots for attention weight distributions of highly weighted fragments and the corresponding contribution value distributions. Right: Fragments with largest and smallest contribution values corresponding to the fragments which tend to decrease and increase the prediction solubility respectively. The average attention weight value for each fragment is shown above each boxplot.}
        \label{fig:logs-weights}
\end{figure*}

\begin{figure*}[t]
    \centering
        \includegraphics[width=.8\textwidth]{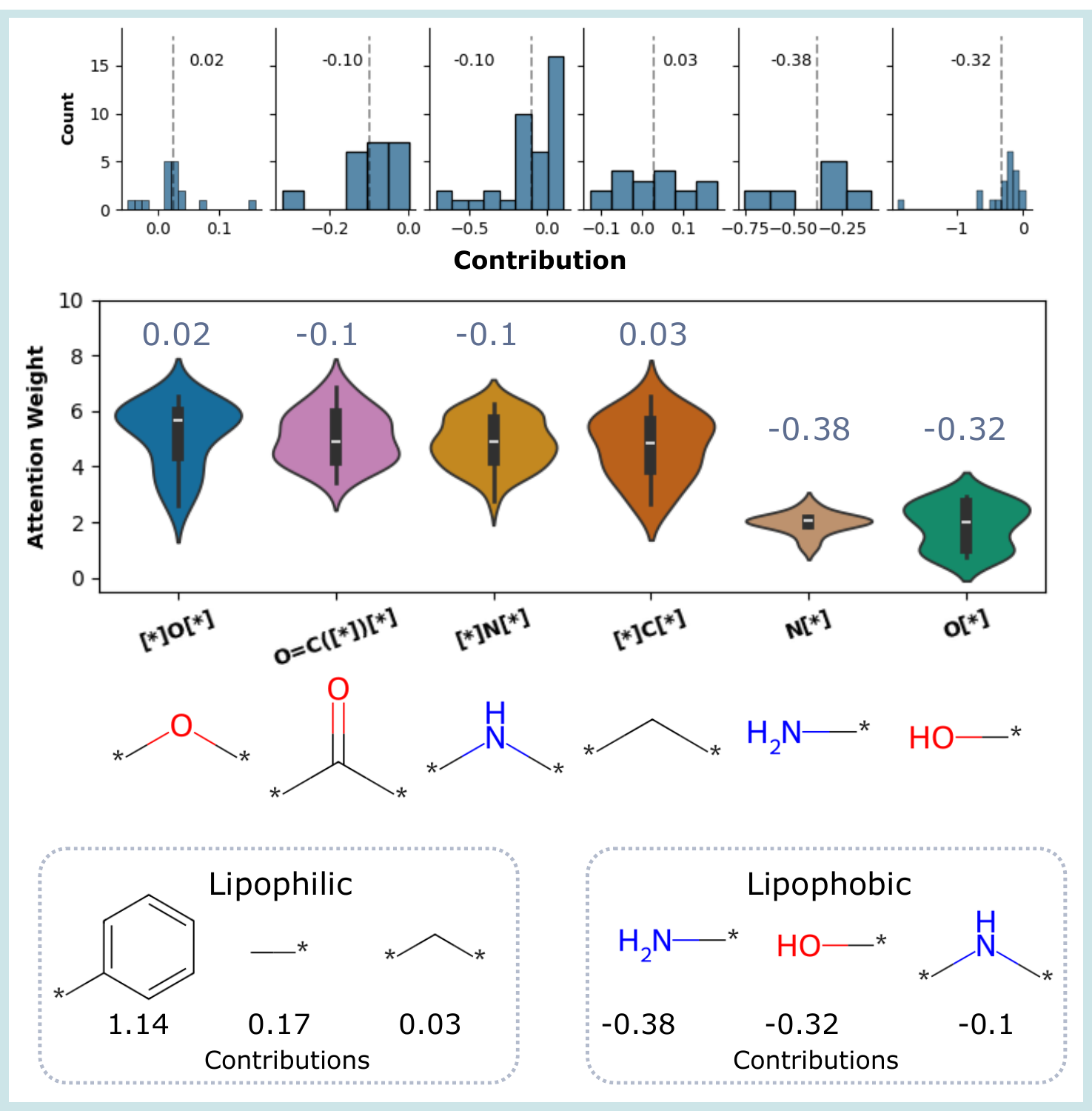}      
        \caption{Fragment and fragment-fragment connection attention weights and contribution values corresponding to lipohilicity prediction. The contribution values corresponding to largest weights are also marked. 
        }
        \label{fig:lipo-weights}
\end{figure*}

\subsubsection*{Atom and Bond Weights}

In Figure \ref{fig:atom-weights-0.1} we illustrate the atoms with the largest average attention weights for solubility, lipophilicity, and cancer drug response (CDR) prediction in molecules with absolute errors less than 0.1. In examining the learned atom‐based importance weights for the three predictive tasks, several salient trends emerge that correspond well with established principles in medicinal and physical chemistry \cite{patrick_introduction_2023}. Heteroatoms—particularly nitrogen, oxygen, and sulfur—consistently receive higher weights across all three models, in line with their known roles in modulating hydrogen bonding, polarity, and reactivity \cite{Hansch1995-lt}. These properties strongly influence molecular solubility and pharmacologically relevant features such as absorption, distribution, metabolism, and excretion (ADME). Carbon, unsurprisingly, also shows high importance in all tasks due to its role as the primary structural scaffold in organic compounds, thereby dictating the molecular framework and overall hydrophobic/hydrophilic balance \cite{jorgensen_many_2004}. Another prominent observation is the relatively higher weighting of chlorine and fluorine for both lipophilicity and cancer‐drug response predictions, consistent with longstanding medicinal chemistry strategies that use halogenation to fine‐tune lipophilicity, metabolic stability, and target engagement \cite{Brown2012-nb}. Finally, hydrogen appears less influential for aqueous solubility, likely because it is overshadowed by the effect of more polar heteroatoms; yet it remains non‐trivial for lipophilicity and biological response, underscoring how subtle alterations in hydrogen content can modulate binding interactions and overall drug‐like behavior \cite{Hansch1995-lt}.



In Figure \ref{fig:bond-weights-0.1},  we show the highly weighted bonds for compounds whose predictions have absolute errors less than 0.1. We notice that C–C and C–N bonds receive consistently high weights for predicting solubility, lipophilicity , and cancer drug response. This trend is unsurprising, as most drug‐like small molecules contain a primarily carbon‐based skeleton alongside strategically placed nitrogen atoms, which can strongly influence pharmacokinetic and pharmacodynamic profiles by modulating hydrogen‐bonding capacity, basicity, and metabolic stability\cite{patrick_introduction_2023}. Notably, bonds that include other heteroatoms, such as C–O or C–S, also appear among the top weights in all three models, underscoring the well‐recognized influence of polar functional groups on aqueous solubility \cite{lipinski_drug-like_2000}. The presence of bonds to halogens (e.g., C–F or C–Cl) shows moderate but still significant influence on cancer drug response prediction, reflecting the common medicinal‐chemistry practice of introducing halogens to tune lipophilicity, improve metabolic stability, and enhance binding affinity \cite{waring_lipophilicity_2010}.

\subsubsection*{Fragment weights}
Next, we study the attention weight and contribution score distributions for the fragment-level substructures. In the left panel of Figure \ref{fig:logs-weights}, we present the attention weight distributions of highly weighted fragments in the context of solubility prediction. Meanwhile, the fragments associated with largest and smallest contribution values are shown in the right panel of Figure \ref{fig:logs-weights}. These results are based on test set predictions with an absolute error below 0.1, and correspond to an $R^{2}$ score of 0.99 between observed and predicted solubility values. 

We find that the fragments containing N are among the high weight fragments ($[*]N([*])[*]$) and high contribution value fragments ($CN(C)[*]$,  $N[*]$).
Nitrogen containing functional groups are known to be hydrophilic. Agreeing with the known chemistry, -OH which is also a well known hydrophilic functional group has on average a positive contribution value.

The compounds containing $[Cl^{-}]$ are salt structures. Our analysis reveals that $Cl^{-}$ is either the most significant or the second most significant fragment in all but one of the compounds containing $Cl^{-}$ in our study set according to the attention weights. Additionally, these compounds contain N\textsuperscript{+}, which ranks among the two most significant substructures in 8 out of the 14 compounds. The fragment weights for these 14 compounds are illustrated in Figure S1. Considering the chemical properties, $Cl^{-}$ is likely to interact predominantly with the positively charged (N) atom. Consequently, the model's identification of these structures as the most important underscores its focus on the relevant substructures.


 Despite the presence of the polar C=O group, the substructures $O=C([*])[*]$ and $O=C([*])C[*]$ also have most of their contribution values between -0.5 and 0. This is likely because of the other fragments attached to these substructures. Upon further inspection, we found that the contribution of $O=C([*])[*]$ is positive when one of the substructures attached to it is OH or NH\textsubscript{2} (see Figure S2). This result is consistent with the well-known fact that the carboxylic acid group (O=C(C)OH) is hydrophilic. When we inspected the compounds where $O=C([*])[*]$ has negative contributions, we find that in majority of them, one end of $O=C([*])[*]$ is attached to a ring system (structures are shown in Figure S2).

\begin{figure*}
    \centering
    \centering
    \includegraphics[width=1\textwidth]{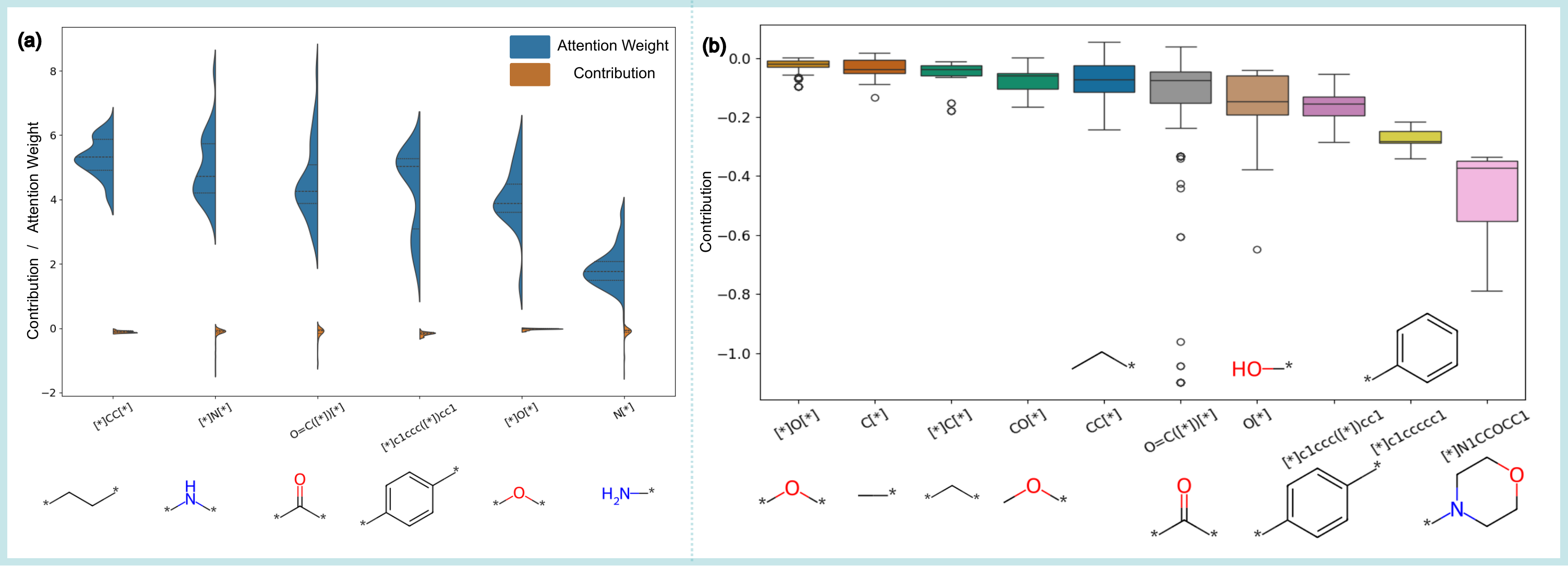}
    \caption{Fragments with (a) largest attention weights for cancer drug response prediction. (b) Fragments corresponding to five largest and five smallest contributions towards CDR prediction.
    } 
     \label{fig:cdrp_frags}
\end{figure*}

\begin{figure*}
    \centering
    \centering
    \includegraphics[width=1\textwidth]{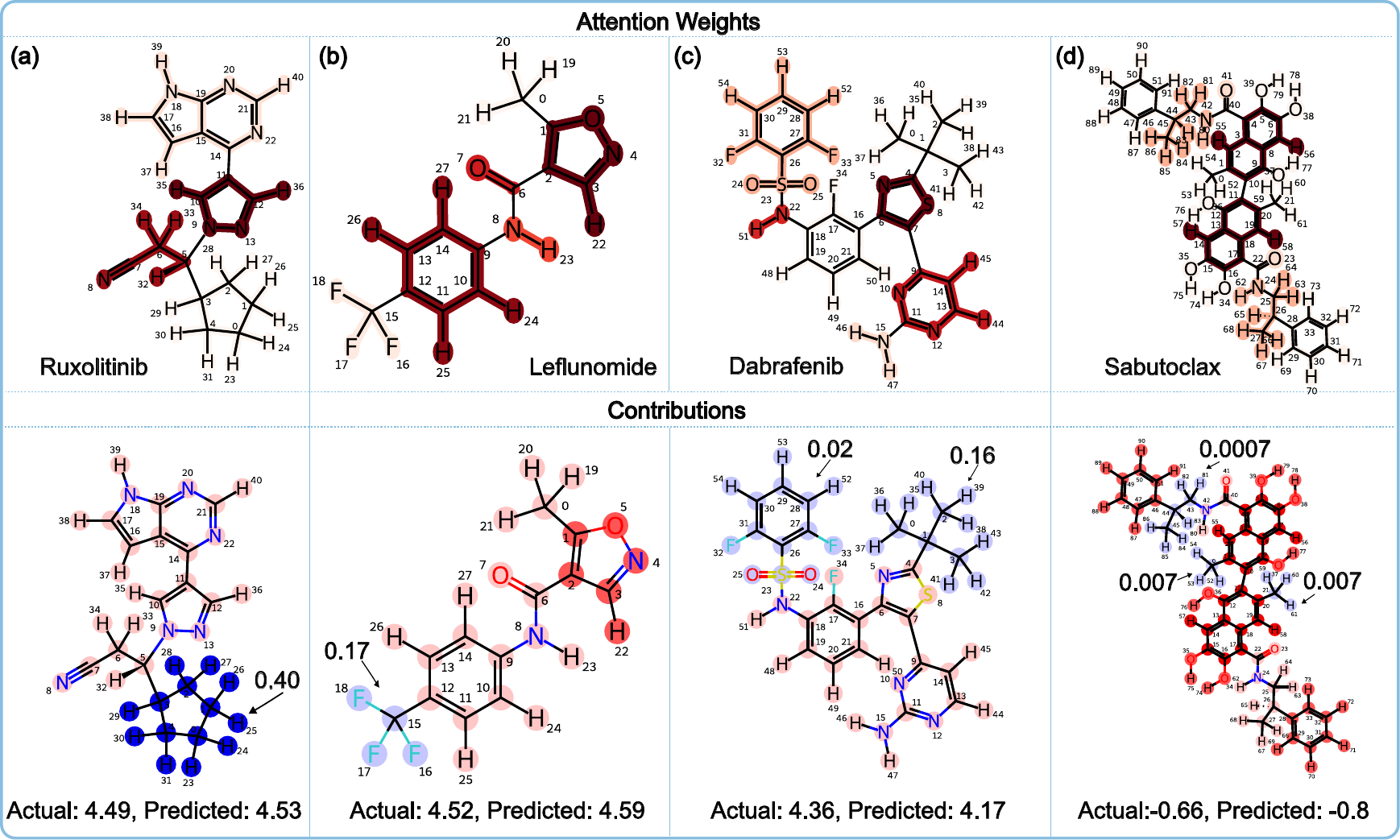}
    \caption{Attention weights (top panel) and contributions (bottom panel) for drugs with large and small drug response values. 
 }
     \label{fig:large_smalll_cdrp}
\end{figure*}



Even though CN(C)[*] has the highest average contribution value, we see that the contributions of N[*] and O[*] can be even larger. As there are only a few such compounds in our study set, we currently cannot distinctly determine the  differences between the compounds where $N[*]$ and $O[*]$ contribute greater than $CN(C)[*]$ for solubility. The substructures containing benzene-like rings are observed to be hydrophobic irrespective of the other structures they are attached to, as their contribution values are consistently negative. As a comparison to solubility, fragment attention weights and contribution values corresponding to lipophilicity are shown in Figure \ref{fig:lipo-weights}. As expected, the trends observed are opposite to those for solubility in terms of the types of substructures which enhance versus diminish the property.




The highly weighted fragments for cancer drug response prediction are shown in Figure \ref{fig:cdrp_frags}(a). Among them are amino groups (–NH\textsubscript{2}, –NR\textsubscript{2}), carbonyl (C=O) groups, ether oxygens, hydroxyl groups and aromatic rings, which are known to be important building blocks of drug molecules~\cite{Thomas2007-nj,Harrold2012-gr}.

Because of their polarity, amino groups increase water solubility and participate in hydrogen bonding and ionic interactions with amino acid residues in proteins, nucleic acids, and other biomolecules. Studies have shown that a compound conjugating with an amino acid results in the improvement of its pharmacological activity\cite{Ma2007-fk,Zhang2008-kp} and water solubility\cite{Drag-Zalesinska2009-wg, Kim2009-op}, and the reduction of its cytotoxicity\cite{Anbharasi2010-js}.



The presence of a carbonyl  (C=O) group can improve the molecular recognition process, allowing the drug to directly fit  into the active sites of the target\cite{1124842806}. Carbonyl groups can enhance the stability of the drug molecule due its ability to form hydrogen bonds. This network of hydrogen bonds helps to prevent premature degradation and ensure that the drug reaches its site of action intact.
The chemical reactivity of carbonyl groups can play a significant role in the mechanism of drug action as well\cite{1124842806}.

\begin{figure*}[t]
    \centering
    \includegraphics[width=1\textwidth]{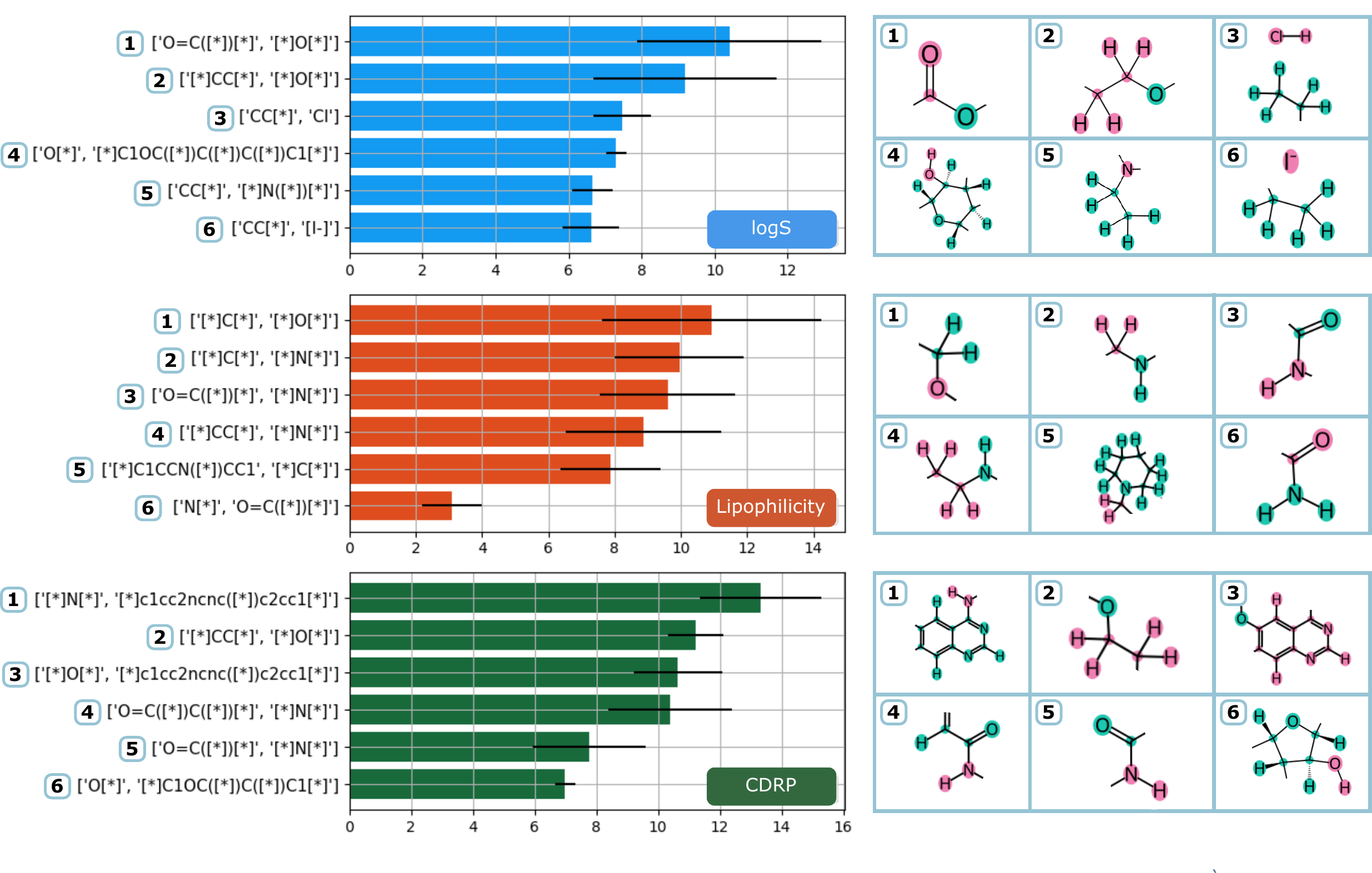}  
    \caption{Highly weighted connections in the contexts of solubility, lipophilicity and cancer drug response prediction.}
    \label{fig:connection_weights}
\end{figure*}


\begin{figure*}[!thb]
    \centering
    \includegraphics[width=1\textwidth]{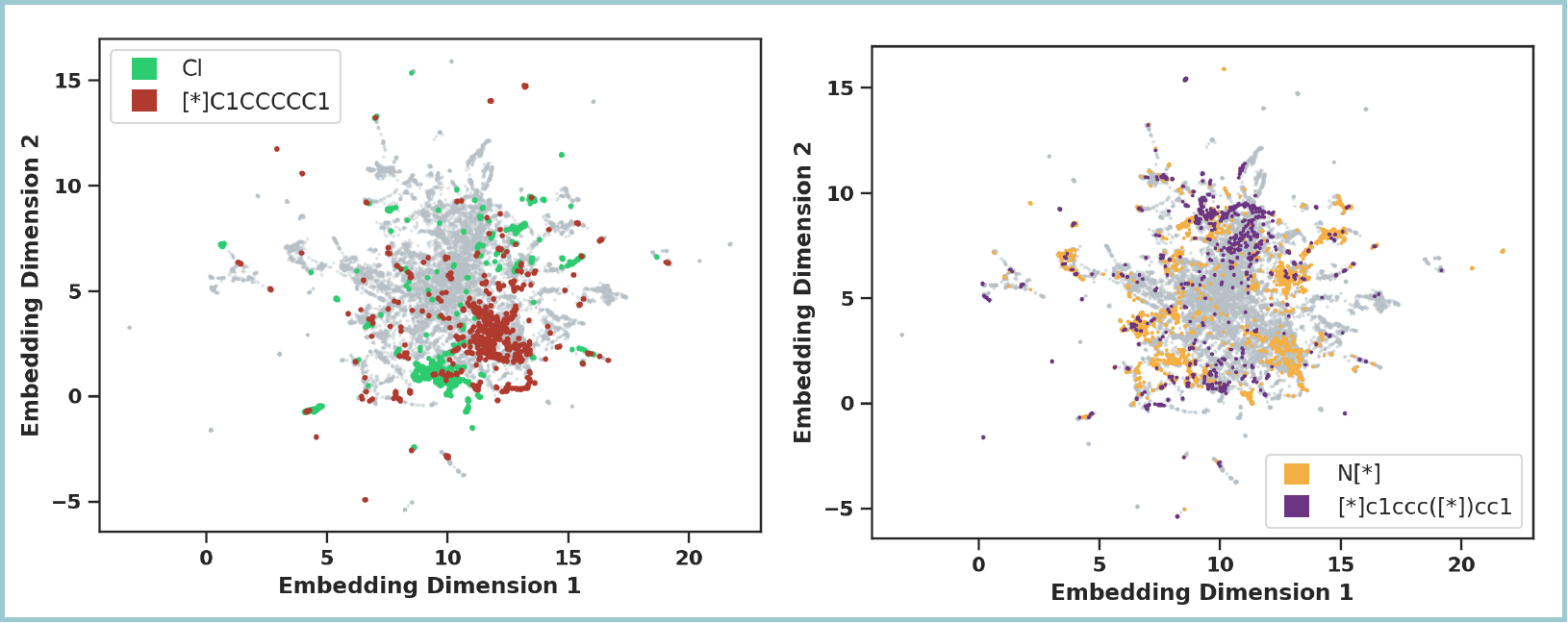}
      
    \caption{2D UMAP projection of FragNet embeddings for molecules in the PNNL solubility dataset highlighting that molecules with hydrophobic and hydrophilic groups occupy different regions of the latent space.}
    \label{fig:umap-logs}
\end{figure*}



Apart from contributing to improve solubility, due to the electron donating and accepting ability of ether oxygens, the presence of ether groups can also increase the bioavailability of a drug\cite{Harrold2012-gr}. 
Ether oxygens (-O-) can also bolster the stability of drug molecules, reducing the chances of degradation before the drug reaches its intended site of action. For instance, in penicillin V, the ether oxygen can attract the lone pair of electrons from the oxygen of the side chain carbonyl group. This withdrawal of electrons prevents the lone pair from attacking the carbonyl carbon of the $\beta$-lactam ring\cite{Harrold2012-gr}.


The presence of hydroyl (-OH) group can increase the solubility and thereby the oral absorption of drug molecules. Hydroxyl groups also affect the metabolism, duration and the selectivity\cite{Harrold2012-gr}. Due to their ability to engage in hydrogen bonding, hydroxyl groups can be employed to modify the sites where hydrogen bonds are formed within a drug molecule. This capability allows for the regulation of both the binding affinity and specificity of the drug towards its biological target\cite{Thomas2007-nj}.


We also find fragments containing aromatic rings among the highly weighted.
Aromatic rings provide hydrophobic interactions and stacking interactions with biological targets and are often present in drugs that need to pass through lipid membranes.

In Figure \ref{fig:cdrp_frags} (b), fragments associated with the five largest and smallest contribution values in our test dataset are shown. Notably, the hydrophobic fragments contribute negatively to the cancer drug response prediction.





To underscore the significance of identifying useful substructures in cancer drug design, Figure \ref{fig:large_smalll_cdrp} presents data on four drugs tested on the same BRCA\cite{Varol2018, Mok_2018} cell line type. Among these, three drugs exhibit high response values (approximately 4.1 to 4.5), while the fourth drug shows a notably lower response value of around -0.66. Analysis reveals that the drugs with high response values have relatively smaller negative contribution values compared to the drug with the low response value. Specifically, the average contributions across fragments for the highly responsive drugs are -0.089, -0.094, and -0.095, while the low response drug has an average contribution of -0.24. Additionally, the highly responsive drugs contain fragments with substantially larger positive contributions compared to the low response drug. The positive contribution values for drugs (a), (b), and (c) are 0.40355, 0.16833, and (0.155, 0.020) respectively, whereas the low response drug shows positive contribution values of just 0.007, 0.007, and 0.001. This indicates that in computational drug design, FragNet's contribution scoring can be utilized to identify initial fragments with positive contributions, thus aiding in the design of more effective cancer drugs.



\begin{figure*}[t]
    \centering
    \includegraphics[width=1\textwidth]{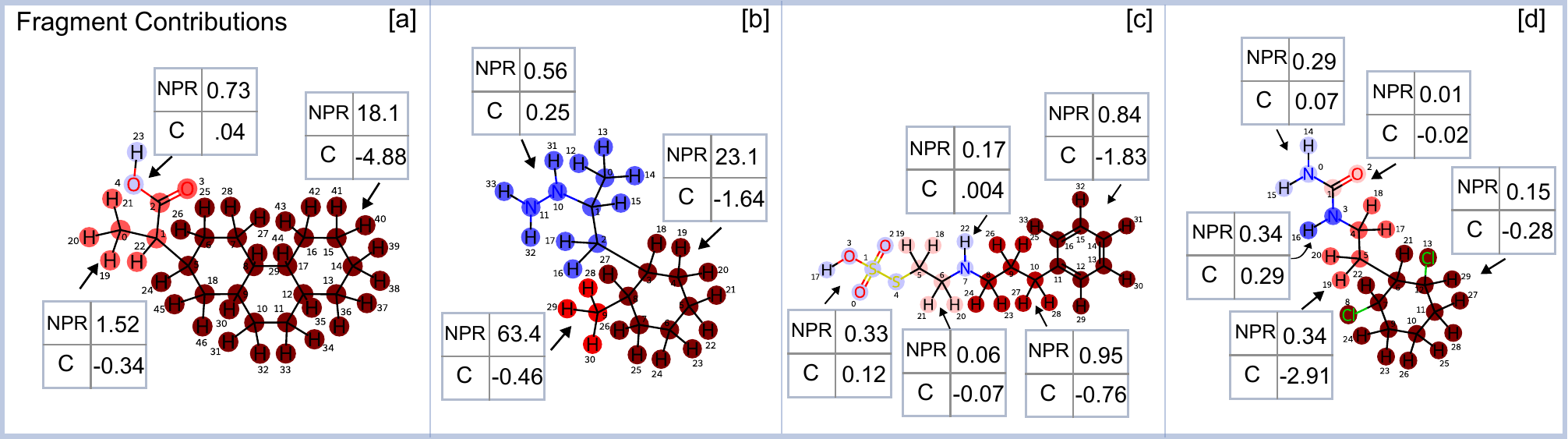}  
    \caption{ Non Polar/Polar ratios (NPR) and contribution (C) values of fragments. It is observed that for highly hydrophobic fragments, the Non Polar Fraction takes large values compared to the hydrophilic fragments.
    }
    \label{fig:dft}
\end{figure*}

\begin{figure}[t]
    \centering
    \includegraphics[width=1\textwidth]{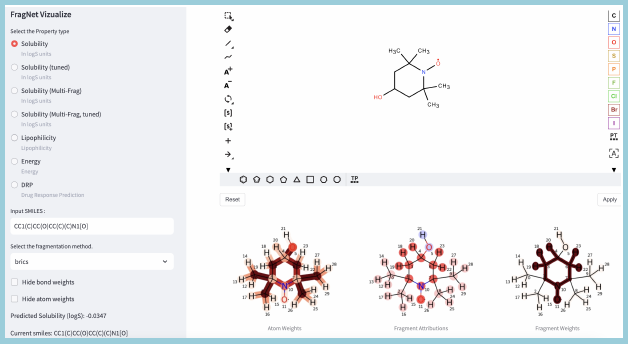}  
    \caption{Interface of the interactive application}
    \label{fig:app}
\end{figure}

\subsubsection*{Fragment-Fragment Connection Weights}

The highly weighted fragment-fragment connections for predicting solubility, lipophilicity, and drug response are shown in Figure \ref{fig:connection_weights}. In most cases, these connections are from oxygen or nitrogen to a carbon-based substituent group. This is unsurprising, given that both oxygen and nitrogen are significantly more electronegative than carbon, producing polar bonds that enable dipole–dipole interactions and, often, hydrogen bonding with polar solvents.

The presence of connections with N-H is notably important for lipophilicity prediction. The covalent bond between a nitrogen atom in an N-H group and a substituent group can significantly alter the electronic environment of the molecule. If the substituent group is electron-withdrawing or electron-donating, it changes the electron density on the nitrogen and the overall polarity of the molecule. This, in turn, affects how the molecule interacts with polar versus nonpolar environments, directly impacting lipophilicity. 
The substituent group can also enable or prevent intramolecular hydrogen bonding or other non-covalent interactions, which can ``hide'' polar groups from the solvent, effectively increasing lipophilicity by reducing exposed polarity.



Additionally, connections with C=O  are also observed to be important for determining all three properties. In particular, the interaction between $[*]CC[*]$ and oxygen is significant for both solubility and drug response predictions. Non-covalent interactions between $CC[*]$ and substructures containing halogen atoms are also noteworthy, as observed in the solubility datasets. Upon close inspection, we found that in almost all cases, these ethyl groups are attached to electron-donating N or O atoms. Therefore, we believe these interactions signify non-covalent interactions between the electronegative group and the anionic substructure. Also, because of the electron donating nature of the ethyl groups, it is possible that there is an attraction between Cl- and slightly positively charged ethyl groups.  This ethyl halogen interaction is also observed for when Cl- is the anion.

\subsection{Additional Analysis}

\subsubsection{Fragments in inaccurately predicted molecules}

To identify molecular fragments that the model finds challenging to predict accurately, we computed the fragment weights and contributions for the top 20\% of molecules exhibiting the largest prediction errors. Subsequently, we categorized these molecules based on their constituent fragments and calculated the average prediction errors, fragment weights, and fragment contributions for each fragment group.

\begin{table*}[!htb]
    \centering
\begin{tabular}{lrrr|rr}
\toprule
Fragment & Error & Weight & Contr. & In low err. & In high err. \\
\midrule
$[*]CC[*]$ & 3.00 & 4.38 & -0.01 & 12.83 & 7.76 \\
$[*]N[*]$ & 2.17 & 4.68 & -0.01 & 13.58 & 15.53 \\
$O[*]$ & 2.13 & 1.61 & 0.06 & 31.32 & 30.12 \\
$[*]N([*])[*]$ & 2.11 & 7.21 & 0.00 & 6.42 & 5.28 \\
$[*]C[*]$ & 1.97 & 4.38 & -0.06 & 3.40 & 8.07 \\
$Cl$ & 1.93 & 6.10 & 0.11 & 8.68 & 4.66 \\
$CO[*]$ & 1.78 & 0.93 & -0.13 & 9.06 & 9.63 \\
$[*]O[*]$ & 1.78 & 4.37 & -0.04 & 28.68 & 20.81 \\
$[Na+]$ & 1.76 & 4.21 & 0.25 & 0.00 & 5.90 \\
$N[*]$ & 1.69 & 1.94 & 0.13 & 8.30 & 11.18 \\
\bottomrule
\end{tabular}
    \caption{Fragments Identified in high-error compounds. `Error' represents the average absolute error; `Weight' denotes the average fragment attention weight; `Contr.' indicates the average fragment contribution to solubility. `In low err.' shows the percentage of low-error molecules containing the fragment, while `In high err.' denotes the percentage of high-error molecules containing the fragment.}
    \label{table:high_err_frag}
\end{table*}

In Table \ref{table:high_err_frag}, we present the fragments found in molecules with high prediction errors, ordered by the magnitude of the errors associated with the molecules they belong to. Specifically, we display the fragments corresponding to the ten largest average errors. To further refine our analysis, we consider fragments with average weight values equal to or above 4. The fragments that meet this criterion include $[*]CC[*]$, $[*]N[*]$, $[*]N([*])[*]$, $[*]C[*]$, $Cl$, $[*]O[*]$, and $[Na+]$.

It is noteworthy that some of these fragments are also present in molecules that were predicted with high accuracy. To identify any fragments predominantly found in high-error molecules, we calculated the percentage of molecules containing each fragment in both low-error and high-error groups. Here, low-error molecules refer to those discussed in Section \ref{substructure_analysis}
 with prediction errors of less than 0.1, while high-error molecules are the top 20\% of molecules with the largest prediction errors considered in this section. These findings are depicted in the right panel of Table \ref{table:high_err_frag}.

From the analysis, we observe that $[Na+]$ is a fragment exclusively found in high-error molecules. Consequently, compounds containing $[Na+]$ can be regarded as challenging for the model to predict accurately. This insight suggests that further analysis is warranted to identify the underlying reasons and potential improvements needed to enhance the model's predictive performance for such molecules.

\subsubsection{Model Embeddings}

To further demonstrate the effectiveness of FragNet's representations in distinguishing different classes of molecules, we present UMAP results in Figure \ref{fig:umap-logs} which show the 2D projection of the embeddings from the layer immediately preceding the final linear layer of the model. The embeddings from this layer should include a condensed representation of the molecular structure that encodes the information necessary for predicting the molecular property. 

We observe that there are distinct regions in the embedding space where molecules with specific fragments are densely populated. This indicates that the learned embeddings for these molecules are significantly influenced by these substructures. Hence, the groupings of molecules in this space provides insight into the reasoning used by the model to infer the target property. For example, with regard to solubility prediction, some molecules containing hydrophilic and hydrophobic fragments are located in separate regions in the embedding space. It should also be noted that the properties of these molecules are not solely determined by a specific functional group attached to them. This is the reason why we notice that molecules containing different functional groups are not in completely isolated clusters.

\subsubsection{Validations using Density Functional Theory (DFT) calculations}

To further investigate the interpretability of FragNet, we calculated the electrostatic surface potentials of some molecules that had the FragNet model trained on solubility predicted with high accuracy using Density Functional Theory (DFT). The details on the DFT calculations are provided in the Supporting Information section. Regions with electrostatic surface potentials (absolute value) greater than 10 were designated as polar areas, while those with potentials less than 10 were classified as non-polar areas. For each molecule, the ratio of the non-polar area to the polar area (NPR) was then calculated for the fragments. A larger NPR indicates a fragment with lower expected polarity. Therefore, we expect fragments with high NPR values to have negative or small contribution values. We observe this trend is in some molecules, as shown in Figure \ref{fig:dft}. This observation is prominent in highly hydrophobic molecules. Specifically, in molecules where the total NPR is greater 1, the average fragment contribution is -1.19, and in molecules where the total NPR is less than 1 the average fragment contribution is -0.35. In Table S1, we provide the fragment contribution and NPR values for 16 molecules.

\subsubsection{Interactive Browser Application}
We are also releasing the code for a web browser-based application that permits users to edit molecules and examine how attention weights vary across different substructures (Figure \ref{fig:app}). This tool is designed for educational purposes and small-scale molecule design. It supports fragmenting a molecule using both BRICS, Murcko, and BRICS+Murcko fragmentation schemes. Currently, users can visualize atom, bond, and fragment weights, as well as fragment contributions. Additionally, substructure weights and substructure-substructure connection weights are displayed.

\section{Future Work}
\begin{figure*}[htb]
    \centering
\includegraphics[width=1\textwidth]{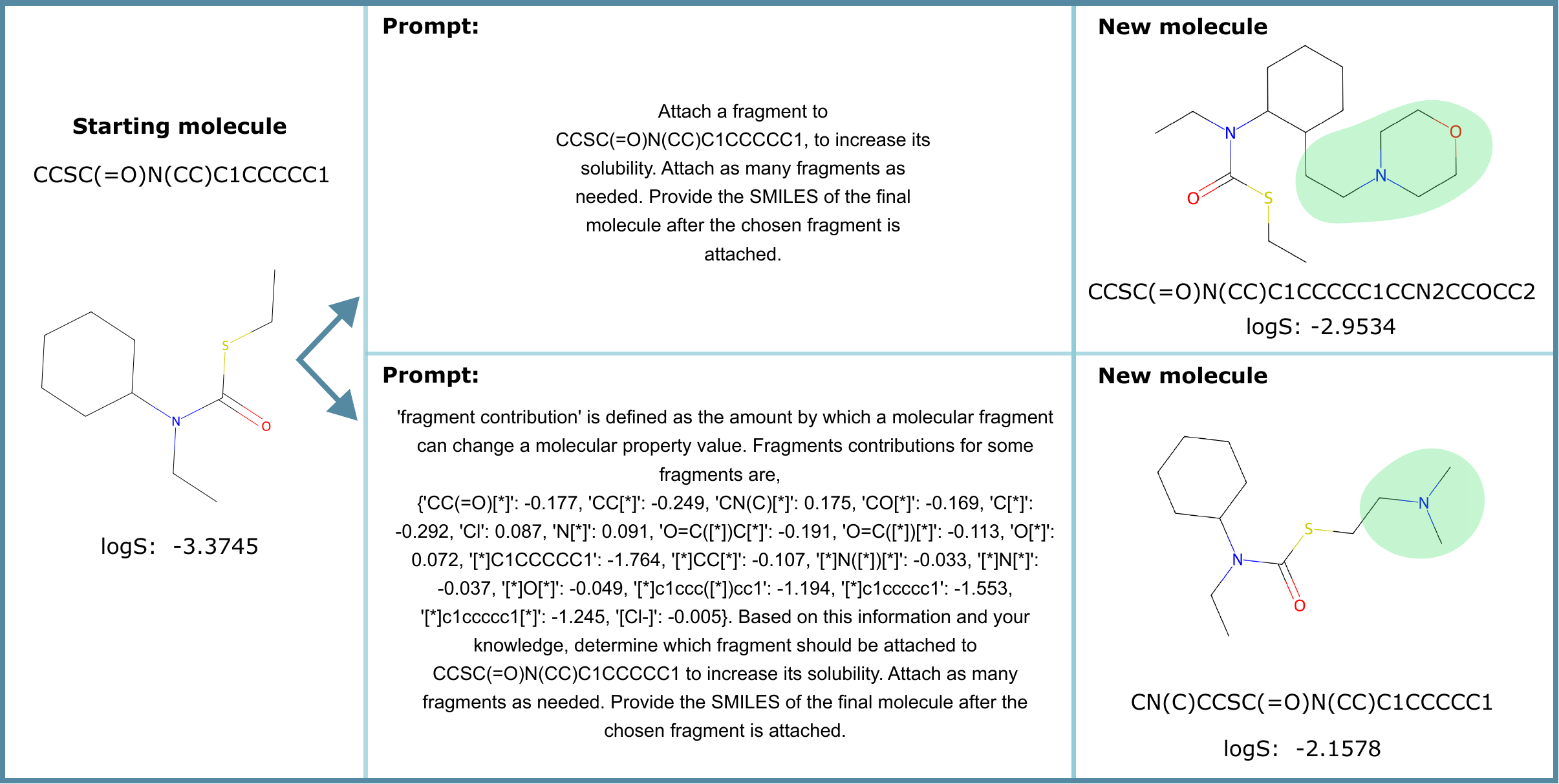}   
    \caption{Prompting Claude 3.5 Sonnet with and without FragNet's contribution data to create a molecule with improved solubility. The new fragments in the LLM-modified molecule are highlighted in green. 
    }
    \label{fig:llm}
\end{figure*}

The recent integration of large language models (LLMs) in explainable AI represents a significant leap towards generating more natural and accessible explanations\cite{gandhi_explaining_2022, wellawatte_human_2025}. LLMs, combined with techniques like Retrieval Augmented Generation (RAG) and tool use, leverage vast repositories of knowledge to provide contextually rich and coherent explanations We envision future developments where FragNet’s interpretability mechanisms are enhanced by integrating LLMs and Vision Language Models to generate explanations that are both scientifically rigorous and easily comprehensible. Furthermore, FragNet can be combined with LLMs for molecular property optimization workflows. In Figure \ref{fig:llm}, we demonstrate how FragNet-predicted contribution values can be utilized to prompt an LLM to design molecules with improved property values.


\section{Conclusion}\label{sec13}
We developed a new graph neural network (GNN) architecture capable of providing model interpretations using atoms, bonds, fragments, and fragment connections. To the best of our knowledge, this is the first GNN architecture that accounts for message passing between non-covalently bonded fragments, thereby creating an improved representation for salts and ionic liquids. The prediction accuracies of our model, FragNet, are comparable to the latest state-of-the-art models. Fragment masking allows for the determination of fragments that enhance or diminish property values, a capability particularly useful in molecular design applications. We find that the model correctly focuses on fragments that most affect solubility, lipophilicity, and drug response based on known chemical properties, thus validating the FragNet architecture.
Our code is publicly available at github/fragnet, and we have also provided an interactive web-based application for editing molecules and visualizing the corresponding attention weights of the substructures.

\section*{Declarations}

\subsubsection*{Funding}
This work was supported by Energy Storage Materials Initiative (ESMI), which is a Laboratory Directed Research and Development Project at Pacific Northwest National Laboratory (PNNL). PNNL is a multiprogram national laboratory operated for the U.S. Department of Energy (DOE) by Battelle Memorial Institute under Contract no. DE- AC05-76RL01830.

\subsubsection*{Conflict of interest}
There are no conflicts of interest to declare
\subsubsection*{Consent for publication}
Not applicable
\subsubsection*{Data availability }
Except for the PNNL solubility dataset, the other datasets are publicly available. The PNNL solubility dataset was created by combining three publicly available datasets. The script to create the PNNL dataset is provided in the code's github repository.
\subsubsection*{Code availability }
The code is available at \url{https://github.com/pnnl/FragNet}
\subsubsection*{Author contribution}
G.P. conceived the idea, led the project, developed the model, analyzed the results and wrote the manuscript. P.G. performed model validation using DFT and composed the corresponding analysis. C.M.M. supervised molecular chemistry analysis of the work and assisted in writing the manuscript. E.G.S. supervised the work, provided project administration, and performed review and editing of the manuscript.
All authors reviewed and approved the final version of the manuscript.

\subsubsection*{Competing interests}
The Authors declare no competing interests.

\begin{acknowledgement}

Please use ``The authors thank \ldots'' rather than ``The
authors would like to thank \ldots''.

The author thanks Mats Dahlgren for version one of \textsf{achemso},
and Donald Arseneau for the code taken from \textsf{cite} to move
citations after punctuation. Many users have provided feedback on the
class, which is reflected in all of the different demonstrations
shown in this document.

\end{acknowledgement}

\begin{suppinfo}


The following files are available free of charge.
\begin{itemize}
  \item Supporting Information - FragNet: A Graph Neural Network for Molecular Property Prediction with Four Levels of Interpretability
\end{itemize}

\end{suppinfo}

\bibliography{sn-bibliography}
\end{document}